\pdfoutput=1

\documentclass[11pt]{article}

\usepackage{acl}

\usepackage{times}
\usepackage{latexsym}

\usepackage[T1]{fontenc}

\usepackage[utf8]{inputenc}

\usepackage{microtype}

\usepackage{amsmath}
\usepackage{amsfonts}
\usepackage{booktabs}
\usepackage{graphicx}
\usepackage{xcolor}
\usepackage{algorithm} 
\usepackage{algpseudocode} 
\usepackage{enumitem}
\usepackage{multirow}
\usepackage{fontawesome}

\title{GOLD: Geometry Problem Solver with Natural Language Description}

\author{Jiaxin Zhang \and ... \and Author n \\
        Address line \\ ... \\ Address line}

\author{Jiaxin Zhang \\
  University of Strathclyde \\
  \texttt{jiaxin.zhang@strath.ac.uk} \\\And
  Yashar Moshfeghi \\
  University of Strathclyde \\
  \texttt{yashar.moshfeghi@strath.ac.uk} \\}

\begin{document}
\maketitle
\begin{abstract}

Addressing the challenge of automated geometry math problem-solving in artificial intelligence (AI) involves understanding multi-modal information and mathematics. \textcolor{black}{Current methods struggle with accurately interpreting geometry diagrams, which hinders effective problem-solving. To tackle this issue, we present the \textbf{G}eometry problem s\textbf{O}lver with natural \textbf{L}anguage \textbf{D}escription (GOLD) model. GOLD enhances the extraction of geometric relations by separately processing symbols and geometric primitives within the diagram. Subsequently, it converts the extracted relations into natural language descriptions, efficiently utilizing large language models to solve geometry math problems.} Experiments show that the GOLD model outperforms the Geoformer model, the previous best method on the UniGeo dataset, by achieving accuracy improvements of 12.7\% and 42.1\% in calculation and proving subsets. Additionally, it surpasses the former best model on the PGPS9K and Geometry3K datasets, PGPSNet, by obtaining accuracy enhancements of 1.8\% and 3.2\%, respectively.\footnote{GOLD code can be found at \url{https://github.com/NeuraSearch/Geometry-Diagram-Description}} 
\end{abstract}

\section{Introduction}

Automated solving of geometry math problems has gained considerable attention in the AI community recently \cite{geoqa, inter-gps, geoqa+, unigeo, pgps9k, GeoDRL, A_Symbolic_Characters_Aware_Model_for_Solving_Geometry_Problems}. Unlike math word problems, geometry math problems involve additional geometry diagrams, necessitating comprehensive reasoning capabilities for understanding multi-modal information (refer to Figure~\ref{fig:model_architecture} for an example of a geometry math problem). As a result, research on automated geometry math problem solving is still in its infancy \cite{unigeo}.

Existing approaches for solving geometry math problems utilize neural networks to embed the diagram and problem text separately or jointly, resulting in highly generalized models \cite{geoqa, unigeo}. However, these methods struggle with accurately capturing the complex relationships within geometry diagrams \cite{math-survey}. Additionally, their vector-based representation of geometric relations is not easily interpretable by humans, posing challenges in identifying whether performance issues are from the relation extraction or the problem-solving component. In a different approach, some studies have successfully translated geometry diagrams into formal languages, enhancing precision and interpretability \cite{symbolic_1, symbolic_2, inter-gps, pgps9k}. \textcolor{black}{However, these methods do not separately process relations among geometric primitives and relations between symbols and geometric primitives, which adds difficulty in solving the geometry math problem correctly. }Moreover, these approaches necessitate specifically designed solvers that take formal languages as input, making them incompatible with prevalent large language models (LLMs).

To address the limitations of existing methods in solving geometry math problems, we introduce the GOLD model. The GOLD model converts geometry diagrams into natural language descriptions, aiding in the generation of solution programs for the problems. Particularly, the GOLD model's relation-construction head extracts two types of geometric relations: \textit{sym2geo} (relations between symbols and geometric primitives) and \textit{geo2geo} (relations among geometric primitives). This process involves two specialized heads that separately model symbols and geometric primitives within diagrams as distinct vectors. These extracted geometric relations are then converted into natural language descriptions. This not only improves the model's interpretability but also connects geometry diagrams with problem texts. Furthermore, since these natural language descriptions meet the input requirements of LLMs, the GOLD model is able to utilize the advanced LLMs as the problem-solving module, efficiently generating solution programs used to solve geometry math problems.

To evaluate the effectiveness of the GOLD model, we conduct experiments on the three latest released datasets: UniGeo (comprising calculation and proving subsets) \cite{unigeo},  PGPS9K \cite{pgps9k}, and Geometry3K \cite{inter-gps}. 
The experimental results show the significant performance gains of our GOLD model compared to state-of-the-art (SOTA) models. It surpasses the Geoformer model, which is the SOTA model on the UniGeo dataset, by 12.7\% and 42.1\% in accuracy on the UniGeo calculation and proving subsets, respectively. Additionally, our GOLD model outperforms the PGPSNet model, the SOTA model on the PGPS9K and Geometry3K datasets by 1.8\% and 3.2\% in accuracy, respectively. These results highlight the superior performance and effectiveness of our proposed GOLD model compared to existing approaches.

The contributions of this work are: \textcolor{black}{(1) We propose the GOLD model to extract geometric relations from geometry diagrams and subsequently convert these relations into natural languages, which are then utilized for solving geometry math problems. Its compatibility with LLMs is a significant advantage, enabling the GOLD model to utilize the capabilities of LLMs to generate solution programs. (2) The GOLD model separately processes symbols and geometric primitives from the diagrams. This separation design simplifies the extraction of the geometric relations.} (3) Our GOLD model demonstrates significant improvements over previous methods across all evaluated datasets, validating the effectiveness of our approach.

\section{Related Work}

Early works have explored solving geometry math problems through rule-based approaches \cite{old_1, old_2, old_3, old_4}. Recently, with the success of deep learning methods, several works have explored using neural network architectures for automated geometry math problem-solving. Approaches such as NGS \cite{geoqa} utilizing LSTM \cite{lstm} and ResNet-101 \cite{ResNet} encoded problem texts and geometry diagrams separately. Later, methods like DPE-NGS \cite{geoqa+} replaced the text encoder with transformer models. However, these methods struggle to effectively integrate problem texts and geometry diagrams. In response, Geoformer \cite{unigeo} emerged, embedding both diagram and problem text jointly using the VL-T5 \cite{VL-T5} model, treating visuals as additional tokens. Despite these advancements, they still struggle to provide precise descriptions of slender, overlapped geometric primitives with complex spatial relationships \cite{pgps5k}, resulting in sub-optimal performance when solving geometry math problems. 

Other approaches typically involve parsing the diagram into formal language and utilizing specific solvers to generate solution programs. Recent works like Inter-GPS \cite{inter-gps} and PGPSNet \cite{pgps9k} employed their parsers to describe the diagram using carefully crafted rules. However, these methods based on predefined rules often lack extensibility, resulting in limited generalization capabilities. To address this issue, our proposed GOLD model generates natural language descriptions of the diagrams, ensuring compatibility of adopting LLMs to generate solution programs.

\section{Model}

\begin{figure*}[tbhp!]
\centering
\includegraphics[width=1.0\textwidth]{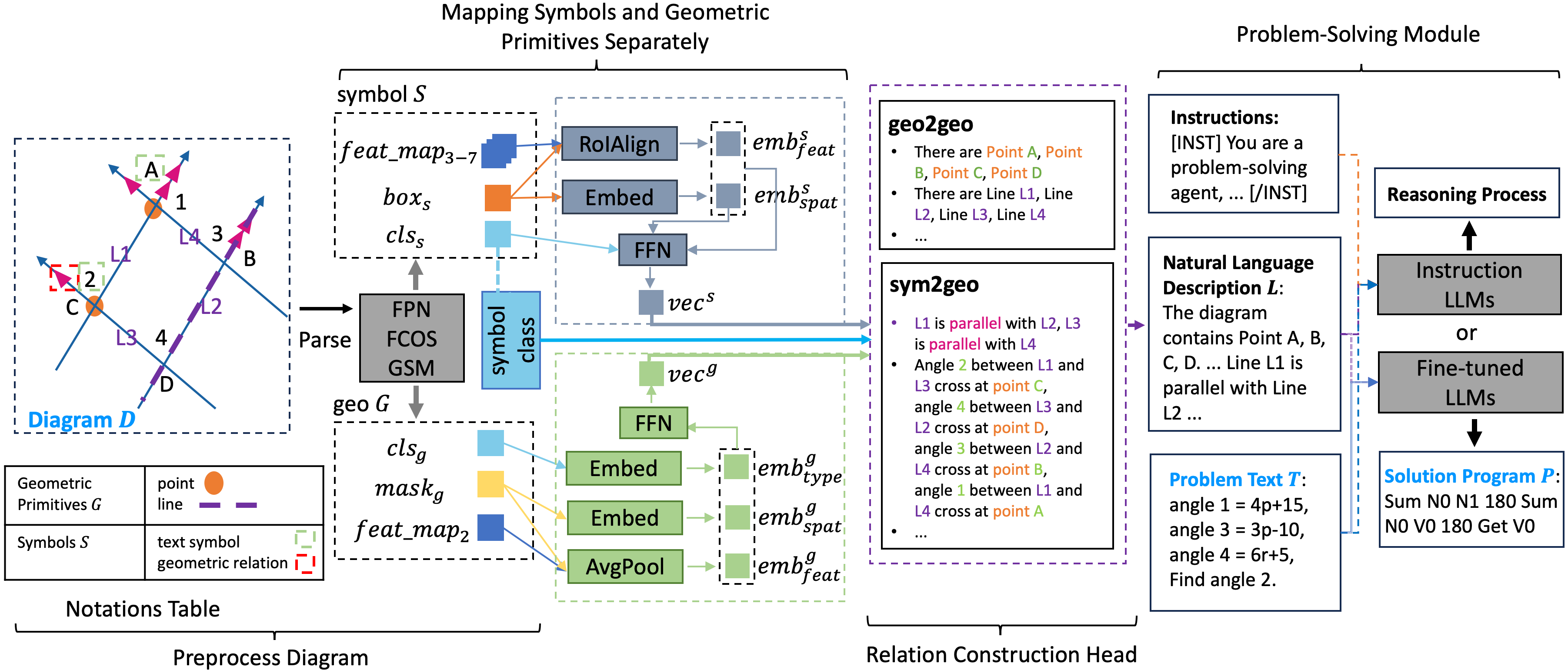}
\caption{The illustration of the GOLD Model. The \textcolor{cyan}{diagram $\mathcal{D}$}, \textcolor{cyan}{problem text $\mathcal{T}$}, and \textcolor{cyan}{solution program $\mathcal{P}$} used in this illustration are sourced from the PGPS9K dataset \citep{pgps9k}. The symbols and geometric primitives in the diagram are annotated using the notations from the Notation Table, which are consistent with the colours of extracted relations of sym2geo and geo2geo.}
\label{fig:model_architecture}
\end{figure*}

Our GOLD model is illustrated in Figure~\ref{fig:model_architecture}.

\subsection{Task Description and Pre-parsing}

The objective is to generate the correct solution program \(\mathcal{P}\) to solve the problem by analyzing a geometry math problem text \(\mathcal{T}\) and its corresponding diagram \(\mathcal{D}\). \textcolor{black}{Specifically, the solution program represents intermediate steps in the domain-specific language generating the output for the question (see an example of solution program in Figure~\ref{fig:model_architecture}).}

In our approach, we initially preprocess geometry diagrams to extract geometric primitives \(\mathcal{G}\) (including \textit{Point} \textbf{P}, \textit{Line} \textbf{L}, and \textit{Circle} \textbf{C}) and symbols \(\mathcal{S}\) from the diagram \(\mathcal{D}\) for subsequent task. Specifically, we utilize a standard Feature Pyramid Network (FPN) \cite{fpn} integrated with a MobileNetV2 \cite{mobilenetv2} backbone for this task. For the detection of symbols, we apply the anchor-free detection model FCOS \cite{fcos}, and for the extraction of geometric primitives, we use the GSM model \cite{pgps5k}. The FCOS model employs feature maps P3 to P7, generated by the FPN layer, to detect symbols within the diagram. This detection step produces bounding box coordinates (\(\text{box}_s\)) and class type (\(\text{cls}_s\)) for each symbol (\(s \in \mathcal{S}\)). For the extraction of geometric primitives, we prefer using the feature map P2 instead of P1, as P2 is more memory-efficient due to its lower resolution. This process results in the identification of segmentation masks \((\text{mask}_g\)) and class type (\(\text{cls}_g\)) for each geometric primitive (\(g \in \mathcal{G}\)).

\subsection{Mapping Symbols and Geometric Primitives Separately}

\textcolor{black}{Before constructing the geometric relations, we map the symbols and geometric primitives into vectors. To achieve this, we introduce two heads: symbol vector head and geometric primitive vector head. Specifically, each head functions as extracting the \textit{feature\_embedding} (\(\text{emb}_{\mathit{feat}}\)) and \textit{spatial\_embedding} (\(\text{emb}_{\mathit{spat}}\)). The \textit{feature\_embedding} is computed from the cropped feature map, which is determined by either the bounding box or the segmentation mask. Moreover, where symbols and geometric primitives are placed significantly shapes how they relate. For instance, only points lying on a line can hold the geometric relation with that particular line. Thus, we hypothesize that incorporating spatial information of \(\mathcal{S}\) and \(\mathcal{G}\) can enhance the accuracy of predictions about geometric relations. Consequently, we embed the bounding boxes of symbols and  the coordinates of the geometric primitives into the \textit{spatial\_embedding}.}

\subsubsection{Constructing the~~\textit{feature\_embedding}}

\textcolor{black}{To obtain the \textit{feature\_embedding} (\(\text{emb}^{s,g}_{\mathit{feat}}\)) and \textit{spatial\_embedding} (\(\text{emb}^{s,g}_{\mathit{spat}}\)) for symbol \(s\) or geometric primitive \(g\), we conduct the below calculation:}

\begin{equation}
\small
	\text{emb}^{s,g}_{\mathit{feat}} = \text{ReLU} (\mathbf{W}^{s,g}_{\mathit{feat}} \mathbf{V}^{s,g})
\label{eq:feature_embedding}
\end{equation}
where \(\mathbf{W}^{s,g}_{\mathit{feat}} \in \mathbb{R}^{h \times h}\) are trainable parameters for either symbols or geometric primitives. Next, we elaborate the calculation process of \(\mathbf{V}^{s,g}\) for symbols and geometric primitives separately.

To obtain the \(\mathbf{V}^{s}\) for symbol \(s\), we utilize RoIAlign \cite{RoIAlign} on its feature map, based on the bounding box of symbol \(s\):

\begin{equation}
\small
	\mathbf{V}^{s} = \mathbb{F} (\text{ReLU} ( \text{BN}( \text{Conv} (\text{RoIAlign} ( \text{box}_s, \text{feat\_map}_{i} )))))
\end{equation}
where \(i\) refers to the \(i\)-th layer of feature maps where the bounding box (\(\text{box}_s\)) is calculated from. The \(\text{Conv}\) is the convolution layer with 64 channels, \(\text{BN}\) is the BatchNorm layer, and \(\text{ReLU}\) is the ReLU activation layer. The \(\mathbb{F}\) means flatten operation, indicating that the \(\mathbf{V}^{s}\) is further flatten into a vector and used for obtaining the \textit{feature\_embedding} \(\text{emb}^{s}_{feat}\) for symbol \(s\) through Eq~\ref{eq:feature_embedding}.

To obtain the \(\mathbf{V}^{g}\) for geometric primitive \(g\), we perform an element-wise multiplication between the segmentation mask \((\text{mask}_{g}\)) of \(g\) and the P2 layer of feature map \((\text{feat\_map}_2\)). Next, we flatten the resulting vector along the height and width dimensions and apply global average pooling to obtain the \(\mathbf{V}^{g}\):
\begin{equation}
\small
	\mathbf{V}^{g} =\text{AvgPool} (\mathbb{F} (\text{mask}_{g} \times \text{feat\_map}_{2}))
\end{equation}
The \(\mathbf{V}^{g}\) is used for calculating the \textit{feature\_embedding} \(\text{emb}^{g}_{feat}\) for geometric primitive \(g\) through Eq~\ref{eq:feature_embedding}.
 
\subsubsection{Constructing the~~\textit{spatial\_embedding}}

The \textit{spatial\_embedding} is obtained by mapping the spatial information of symbols and geometric primitives into embeddings. Specifically, for symbol \(s\), we map the coordinates of its bounding box into an embedding using the trainable parameters \(\mathbf{W}^{s}_{\mathit{spat}} \in \mathbb{R}^{h \times 4}\). Specifically, \(\text{emb}^{s}_{\mathit{spat}} \mathbin{=} \mathbf{W}^{s}_{\mathit{spat}}[x_t, y_t, x_b, y_b]^\top\), where \((x_t, y_t)\) represent the coordinates of the top-left corner of the bounding box, and \((x_b, y_b)\) is the coordinates of the bottom-right corner of the bounding box.

Next, to obtain the \textit{spatial\_embedding} of a geometric primitive \(g\), we start by representing coordinates of \(g\) using \(loc_{g}\). The format of \(loc_{g}\) depends on the class type (\(\text{cls}_{g}\)) of the geometric primitive: for a point, it contains two numbers (\(n_g=2\)) representing its coordinates; for a line, it contains four numbers (\(n_g=4\)) representing the coordinates of its start and end points; and for a circle, it contains three numbers (\(n_g=3\)) representing the coordinates of its centre point and the radius length. We then map \(loc_{g}\) into \textit{spatial\_embedding} by calculating \(\text{emb}^{g}_{\mathit{spat}} = \text{ReLU} (\mathbf{W}^{g}_{\mathit{spat}} (\mathbf{W}^{g}_{\mathit{loc}} loc_{g}\))), where \(\mathbf{W}^{g}_{\mathit{loc}} \in \mathbb{R}^{h \times n_g}\) are different trainable parameters for different \(\text{cls}_{g}\), and  \(\mathbf{W}^{g}_{\mathit{spat}} \in \mathbb{R}^{h \times h}\) are trainable parameters. 

To help the model differentiate between different types of geometric primitives, we introduce the \textit{geo\_type\_embedding} (\(\text{emb}^{g}_{\mathit{type}}\)) to capture the semantic information of the geometric primitive.  The \(\text{emb}^{g}_{\mathit{type}}\) is obtained by performing a lookup operation on the embeddings using the class type \((\text{cls}_{g}\)) of \(g\) from the list of geometric primitive types \([\mathbf{P}, \mathbf{L}, \mathbf{C}]\). Specifically, \(\text{emb}^{g}_{\mathit{type}} \mathbin{=} \text{embedding}(\text{cls}_{g})\), where \(\text{cls}_{g}\) is the class type ID of \(g\). 

\subsubsection{Symbol Vector and Geometric Primitive Vector}

\textcolor{black}{The vector representation \(\text{vec}^{s \in {\mathcal{S}} }\) of symbol \(s\) is obtained by passing concatenated \(\text{emb}^{s}_{\textit{feat}}\) and \(\text{emb}^{s}_{\textit{spat}}\) through a specific feed-forward neural network:}

\begin{equation}
\small
	\text{vec}^{s \in {\mathcal{S}} } = \text{ReLU} ( \mathbf{W}^{s}_{\mathit{vec}} [\text{emb}^{s}_{\mathit{feat}} : \text{emb}^{s}_{\mathit{spat}}]^\top)
\label{eq:symbol_vector}
\end{equation}
where \(\mathbf{W}^{s}_{\mathit{vec}} \in \mathbb{R}^{h \times 2h}\) are the trainable parameters depending on the class type (\(\text{cls}_s\)) of symbol \(s\), and \([ : ]\) refers to concatenation operation.

\textcolor{black}{The vector representation of the geometric primitive \(\text{vec}^{g \in {\mathcal{G}} }\) is obtained by summing up three embeddings relevant to the geometric primitive \(g\), \(\text{emb}^{g}_{\mathit{feat}}\), \(\text{emb}^{g}_{\mathit{spat}}\), and \(\text{emb}^{g}_{\mathit{type}}\):}

\begin{equation}
\small
	\text{vec}^{g \in {\mathcal{G}} } = \text{ReLU} (\mathbf{W}^{g}_{\mathit{vec}} ( \text{emb}^{g}_{\mathit{feat}} + \text{emb}^{g}_{\mathit{spat}} + \text{emb}^{g}_{\mathit{type}}))
\label{eq:geometry_primitive_vector}
\end{equation}
where \(\mathbf{W}^{g}_{\mathit{vec}} \in \mathbb{R}^{h \times h}\) are the trainable parameters.

\subsection{Relation Construction Head}
The relation-construction head aims to establish \textit{sym2geo} relations among symbols and geometric primitives and \textit{geo2geo} relations among geometric primitives themselves.

\subsubsection{sym2geo relation}

The \textit{sym2geo} relation can be further divided into \textit{text2geo} and \textit{other2geo} relations. The \textit{text2geo} relation explains the association between text symbols and geometric primitives, where the text symbols are used to be the reference to a geometric primitive or to display degree, length, etc. To distinguish the role of a text symbol, we introduce the \textit{text\_class} for the text symbol. Specifically, when \textit{text\_class} is category \(\mathit{0}\), the \textit{text2geo} signifies point (or line, or circle) names; when \textit{text\_class} is category \(\mathit{1}\), the \textit{text2geo} corresponds to angle degrees; when \textit{text\_class} is category \(\mathit{2}\), the \textit{text2geo} signifies line lengths; when \textit{text\_class} is category \(\mathit{3}\), the \textit{text2geo} denotes the degree of an angle within a circle. The probabilities of the category (\(P(\textit{text\_class}|s)\)) of text symbol (\(s \in \{\mathcal{S} | \text{cls}_s = \text{"text"}\}\)) is defined as:

\begin{equation}  
\small
 P = \text{softmax} (\mathbf{W}^{\mathit{sym2geo}}_{\mathit{text\_class}} \text{ReLU} (\mathbf{W}^{\mathit{sym2geo}}_{1} \text{vec}^{s}) )
\label{eq:text-class}
\end{equation}
where \(\mathbf{W}^{\mathit{sym2geo}}_{1} \in \mathbb{R}^{h \times h}\) and  \(\mathbf{W}^{\mathit{sym2geo}}_{\mathit{text\_class}} \in \mathbb{R}^{4 \times h}\), both are the trainable parameters. 

The \textit{other2geo} relation captures relations between non-text symbols (\(s \in \{\mathcal{S} |\text{cls}_s \neq \text{"text"}\}\)) and geometric primitives. \textcolor{black}{The non-text symbols are used to find out the relations among geometric primitives, such as \textit{angles of same degree}, \textit{lines of same length}, \textit{parallel lines}, and \textit{perpendicular lines}. For instance, in Figure~\ref{fig:model_architecture}, the symbol enclosed in a red rectangle signifies the parallel relation.}

To establish the \textit{sym2geo} relation between symbol \(s\) and geometric primitive \(g\), we begin by utilizing the corresponding symbol head to transform the vector of the geometric primitive: \(\widehat{\text{vec}}^{g} \mathbin{=} \text{ReLU} (\mathbf{W}^{\mathit{sym2geo}}_{s1} \text{vec}^{g})\), 
where \(\mathbf{W}^{\mathit{sym2geo}}_{s1} \in \mathbb{R}^{h \times h}\) are trainable parameters that vary depending on different class types  (\(\text{cls}_s\)) of symbols. Finally, we calculate the probabilities of the existence of the relation between symbol \(s\) and geometric primitive \(g\) as follows:

\begin{equation}  
\small
\begin{aligned}
 & O_1 \mathbin{=} \text{ReLU}( \mathbf{W}^{\mathit{sym2geo}}_{2} [\text{vec}^{s} : \widehat{\text{vec}}^{g \in \{sub\}}]) \\ 
 & P(\text{rel}^{\mathit{sym2geo}}_{s,g}|s,g) \mathbin{=} \text{sigmoid} (\mathbf{W}^{\mathit{sym2geo}}_{rel}  O_1)
\end{aligned}
\end{equation}
where \(\mathbf{W}^{\mathit{sym2geo}}_{2} \in \mathbb{R}^{h \times 2h}\) and \(\mathbf{W}^{\mathit{sym2geo}}_{rel} \in \mathbb{R}^{1 \times h}\) are the trainable parameters. Worth mentioning, that each type of symbol, including the additional four categories of the text symbol, has its own \(\mathbf{W}^{\mathit{sym2geo}}_{2}\). Additionally, \(\{sub\}\) refers to the subset of geometric primitives, as certain symbols can only have relations with specific geometric primitives. Please refer to Appendix~\ref{sec:appendix_Inference_sym2geo} for details on how to predict \textit{text2geo} and \textit{other2geo} relations during the inference stage. 



\subsubsection{geo2geo relation}

\textcolor{black}{Previous work tend to provide only \textit{sym2geo} relations. However, despite the \textit{sym2geo} relation can provide geometric relations among geometric primitives like parallel, perpendicular, etc. We hypothesize that providing additional information that describes all the geometric primitives from the diagrams is beneficial for the task.} Moreover, we tackle the issue concerning the absence of references to geometric primitives in the diagram. For example, in Figure~\ref{fig:model_architecture}, the original diagram lacks a reference to the line, where \textit{sym2geo} relation cannot address. To overcome this limitation, we have devised an automated approach that assigns appropriate references to the geometric primitives using the format "\(\text{cls}_g\) + num" (e.g., "L1, L2, L3, L4" in purple in Figure~\ref{fig:model_architecture}). This enables the relation-construction module to (1) present a detailed depiction of the diagram by describing the \textit{geo2geo} relations, even in the absence of a single reference, and (2) generate all \textit{sym2geo} relations, even when some geometric primitives lack references. The \textit{geo2geo} relations are categorized according to the involved geometric primitives: (1) Point and Line: "on-a-line" and "end-point". The "on-a-line" relation occurs when a point lies between the tail and the head of the line. Specifically, a point lying at either the head or the tail of the line is the "end-point", which is the special case of "on-a-line". (2) Point and Circle: "centre-point" and "on-a-circle." The "centre-point" relation refers to a point being the centre point of the circle. The "on-a-circle" relation occurs when a point lies on the arc of the circle. Finally, the probabilities (\(P(\text{rel}^{\mathit{geo2geo}}_{g_i,g_j}|g_i,g_j)\)) of the relations between geometric primitives \(g_i\) and \(g_j\) can be calculated as follows:

\begin{equation}  
\small
 P = \text{softmax} (\mathbf{W}^{\mathit{geo2geo}}_{\mathit{rel}} \text{ReLU}( \mathbf{W}^{\mathit{geo2geo}}_{1} (\text{vec}^{g_i} + \text{vec}^{g_j})))
\label{eq:geo2geo-relation}
\end{equation}
where \(\mathbf{W}^{\mathit{geo2geo}}_{1} \in \mathbb{R}^{h \times h}\) and  \(\mathbf{W}^{\mathit{geo2geo}}_{\mathit{rel}} \in \mathbb{R}^{3 \times h}\) are the trainable parameters (the number 3 refers to "no relation" and two relations from either Point and Line or Point and Circle). Please refer to Appendix~\ref{sec:appendix_Inference_geo2geo} for details on how to predict \textit{geo2geo} relations during the inference stage.



\subsection{Problem-Solving Module}

Both the \textit{sym2geo} and \textit{geo2geo} relations are expressed in natural languages by the GOLD model, following the same format as the problem text \(\mathcal{T}\) (please refer to Appendix~\ref{sec:appendix_convert} for the paradigm of converting \textit{sym2geo} and \textit{geo2geo} relations to natural language descriptions). Therefore, it is convenient to utilize the LLMs as the problem-solving module. Specifically, the problem text \(\mathcal{T}\) and the natural language descriptions \(\mathcal{L}\) are concatenated for the LLMs to generate the solution program \(\mathcal{P}\). \textcolor{black}{To illustrate the compatibility of our methods with LLMs, we employ three well-known models for problem-solving: T5-base \cite{T5}, Llama2-13b-chat \cite{llama2}, and CodeLlama-13b \cite{codellama}. The T5-base model is fine-tuned for the target solution programs. Conversely, for Llama2-13b-chat and CodeLlama-13b, we employ directive instructions to guide their solution generation process (please refer to Appendix~\ref{sec:appendix_instruciton_choice} for the choice of instructions).}

\subsection{Training Objective}

Given a dataset of geometry math problems. The training process begins with training the pre-parsing module to extract necessary features from the geometry diagrams. Following this, we focus on training three components: the symbol vector head, the geometric primitive vector head, and the relation-construction head. \textcolor{black}{This training is guided by minimizing a joint loss function, which is defined as \(\mathit{L}_{\mathit{cons}} \mathbin{=} \mathit{L}_{\mathit{g2g}} \mathbin{+} \mathit{L}_{\mathit{t\_cls}} \mathbin{+} \mathit{L}_{\mathit{s2g}}\). The \(\mathit{L}_{\mathit{g2g}}\) loss represents the negative log-likelihood loss for accurately identifying the ground truth \textit{geo2geo} relations. Meanwhile, the \(\mathit{L}_{\mathit{t\_cls}}\) constitutes the negative log-likelihood loss for correctly categorizing the text symbols. Lastly, the \(\mathit{L}_{\mathit{s2g}}\) loss is the binary cross-entropy loss associated with the ground truth \textit{sym2geo} relations. Once they are trained, and their parameters are fixed, we advance to the final stage of fine-tuning the problem-solving module.\footnote{Note that the fine-tuning step is only implemented when T5-base is used as the problem-solving module.} During this stage, our objective is to minimize the \(\mathit{L}_{\mathit{prog}}\) loss, which is the negative log-likelihood loss for correct solution programs} (please refer to Appendix~\ref{sec:appendix-loss-function-details} for more details of loss functions).

\section{Experiments and Results}
\label{sec:experiments}

\begin{table*}[tbht!]
\begin{small}
\centering
\begin{tabular}{@{}lcccc@{}}
\toprule
Models & \multicolumn{1}{c}{UniGeo CAL Test (\%)} & \multicolumn{1}{c}{UniGeo Prv Test (\%)} & \multicolumn{1}{c}{PGPS9K Test (\%)} & \multicolumn{1}{c}{Geometry3K Test (\%)} \\  \midrule 
BERT2Prog     & 54.7$\dagger$ & 48.0$\dagger$  &  -    &  -     \\
NGS           & 56.9$\dagger$ & 53.2$\dagger$  & 34.1$\ddagger$ & 35.3$\ddagger$  \\
Geoformer     & 62.5$\dagger$ & 56.4$\dagger$  & 35.6$\ddagger$ & 36.8$\ddagger$  \\ \midrule
InterGPS      & 56.8 & 47.2  & 38.3 & 48.6  \\
InterGPS (GT) & n/a  & n/a   & 59.8$\ddagger$ & 64.2$\ddagger$  \\
PGPSNet       & 53.2 & 42.3  & 58.8 & 59.5  \\
PGPSNet (GT)  & n/a  & n/a   & 62.7$\ddagger$ & 65.0$\ddagger$    \\ \midrule
GOLD          & \textbf{75.2}  & \textbf{98.5}   & \textbf{60.6} & \textbf{62.7} \\
GOLD (GT)     & n/a  & n/a   & 65.8  & 69.1 \\ \bottomrule
\end{tabular}
\caption{Comparison results on the test subsets of chosen datasets. PGPSNet reported models' performances using the ground truth diagram annotations, where these models have "(GT)" behind them. We re-implemented these methods to get performances without GT annotations. Note that UniGeo lacks GT diagram annotations, so relevant cells are "n/a". "$\dagger$" and "$\ddagger$" indicates the results are from \citealp{unigeo} and \citealp{pgps9k}, respectively.}
\label{tab:overall_result}
\end{small}
\end{table*}

\subsection{Experimental Setup}

Our method was implemented using the PyTorch \cite{pytorch} and HuggingFace \cite{huggingface} libraries. For the pre-parsing module, we followed the training and parameter settings of the previous work \cite{pgps5k}. \textcolor{black}{We evaluated the dimensions of the embeddings over a range of \{32, 64, 128\}, and based on the model's performance in the validation set, we experimentally determined 64 as the optimal dimension size for the embeddings.} We utilized the Adam optimizer with a learning rate of \(1e^{-4}\) and weight decay of \(1e^{-4}\) for training all modules. The symbol vector head, geometric primitive vector head, and relation-construction head were trained end-to-end for 50 epochs with a batch size of 20, while the problem-solving module (using T5-base) was fine-tuned for 30 epochs with a batch size of 10. All experiments were conducted on one NVIDIA A100 80GB GPU.

\subsection{Datasets}

Our experiments are conducted on three datasets: UniGeo \cite{unigeo}, PGPS9K \cite{pgps9k}, and Geometry3K \cite{inter-gps}. The UniGeo dataset comprises 14,541 problems, categorized into 4,998 calculation problems (CAL) and 9,543 proving problems (PRV), \textcolor{black}{which are split into train, validate, and test subsets in a ratio of 7.0: 1.5: 1.5. The Geometry3K includes 3,002 problems, divided into train, validate, and test subsets following a 7.0: 1.0: 2.0 ratio. Since PGPS9K contains a partial Geometry3K dataset, we keep an exclusive set of 6,131 problems, of which 1000 problems are a test subset. Due to the absence of a validation subset in PGPS9K, we divide its training set to create a train-validation split in a 9.0: 1.0 ratio.}

\subsection{Evaluation Metrics}

To compare against existing works, we adhere to the evaluation criteria from the original datasets for both our model and the baselines. For the UniGeo dataset, we utilize the top-10 accuracy metric, which measures the ratio of correct solution programs among the top ten predictions, aligning with the metric used by the authors of the UniGeo dataset. For the PGPS9K and Geometry3K datasets, we adopt a stricter metric, the top-3 accuracy, as recommended by the authors of the PGPS9K dataset. Note that our comparison involves matching the predicted solution program with the ground truth, which is more rigorous than merely comparing the numerical output derived from the solution program.\footnote{\textcolor{black}{This is grounded in the principle that a correct output can sometimes be produced by an incorrect solution program, indicating a failure in the model's understanding of the problem. For example, consider a problem where the correct answer is "5" and the correct program is "2 × 3 - 1". An incorrect program like "2 + 3" could still yield the correct output. Thus, generating the correct program is a more reliable indicator of the model's accurate problem comprehension.}}

\subsection{Comparison with State-of-the-art Models}

We evaluate the performance of our GOLD model (\textit{using T5-base as its problem-solving module}) against state-of-the-art (SOTA) methods in solving geometry math problems. The selected baselines for this comparison include: 1. \textbf{PGSPNet} \cite{pgps9k}: it integrates a combination of CNN and GRU encoders, which generate an encoded vector of the diagram that serves as the input aligning with the logic form to the solver module. 2. \textbf{Inter-GPS} \cite{inter-gps}: it parses both the problem text and the diagram into a formal language, subsequently feeding this into the solver. 3. \textbf{Geoformer} \cite{unigeo}: it utilizes the VL-T5 model for the purpose of diagram encoding, then servers encoded embeddings to the transformer. 4. \textbf{NGS} \cite{geoqa}: it uses the ResNet-101 for its encoding process, showcasing a different approach in handling the diagram encoding. 5. \textbf{Bert2Prog} \cite{geoqa}: it leverages BERT and ResNet as encoders and an LSTM network for generating.

The results presented in Table~\ref{tab:overall_result} demonstrate that our GOLD model outperforms baselines across test subsets of all datasets. Specifically, when compared to Geoformer, the SOTA on the UniGeo dataset, our model exhibits a remarkable increase in accuracy: 12.7\% on the UniGeo CAL and 42.1\% on the UniGeo PRV. Compared to the SOTA model on PGPS9K and Geometry3K datasets, PGPSNet, the GOLD model surpasses it by 1.8\% and 3.2\% in accuracy, respectively. When using ground truth diagram annotations, the GOLD (GT) shows a significant improvement in accuracy on the PGPS9K and Geometry3K, with gains of 3.1\% and 4.1\% over PGPSNet (GT). Against InterGPS (GT), the improvements are at 6.0\% and 4.9\%, respectively. These results underline the effectiveness of the GOLD model in solving geometry math problems.

Moreover, our GOLD model distinguishes itself from approaches like InterGPS and PGPSNet, which rely on logic-form representations to describe diagrams. In contrast, GOLD inputs natural language descriptions to LLMs to generate solution programs. Using natural language leads to significant improvements across all datasets compared to InterGPS and PGPSNet, as evidenced in Table~\ref{tab:overall_result}. Furthermore, models like Geoformer and NGS primarily encode diagrams into vectors. These approaches fall short in providing precise descriptions of the diagrams and limit the adoption of LLMs, thus leading to worse performances compared to our GOLD model. This highlights the importance of detailed and accurate diagram representations for tackling geometry math problems, where our GOLD model excels.

Worth mentioning is that the training for the symbol vector head, geometric primitive vector head, and relation-construction head of the GOLD model was exclusively conducted on the PGPS9K and Geometry datasets due to the lack of annotations in the UniGeo dataset. Despite this, the outstanding performance of the GOLD model on the test subset of UniGeo, as shown in Table~\ref{tab:overall_result}, demonstrates its exceptional generalization capability.

\subsection{Ablation Study on Natural Language Description}
\label{sec:Ablation-Study-on-Natural-Language-Description}

\textcolor{black}{We assess our model's efficacy using three distinct diagram description formats: absence of diagram description, logic forms, and natural language descriptions. The comparative results are detailed in Table~\ref{tab:nldvslf}. When fine-tuning T5-base as the problem-solving module, Table~\ref{tab:nldvslf} indicates that descriptions in natural language outperform those in logic-form, with 3.1\% and 3.4\% improvements on the test subsets of PGPS9K and Geometry3K, respectively.}

\begin{table}[tbhp!]
\centering
\begin{scriptsize}
\begin{tabular}{@{}lcccccc@{}}
\toprule
 & \multicolumn{3}{c}{PGPS9K} & \multicolumn{3}{c}{Geometry3K} \\ \cmidrule(lr){2-4} \cmidrule(lr){5-7}
 & n/a      & LF     & NLD     & n/a       & LF       & NLD      \\ \midrule
T5-base &
  \begin{tabular}[c]{@{}c@{}}22.3\\ $\pm$ \textit{0.0}\end{tabular} &
  \begin{tabular}[c]{@{}c@{}}57.5\\ $\pm$ \textit{0.3}\end{tabular} &
  \begin{tabular}[c]{@{}c@{}}\textbf{60.6}\\ $\pm$ \textit{0.3}\end{tabular} &
  \begin{tabular}[c]{@{}c@{}}12.3\\ $\pm$ \textit{0.0}\end{tabular} &
  \begin{tabular}[c]{@{}c@{}}59.3\\ $\pm$ \textit{0.5}\end{tabular} &
  \begin{tabular}[c]{@{}c@{}}\textbf{62.7}\\ $\pm$ \textit{0.2}\end{tabular} \\ \midrule
Llama2-13b-chat &
  \begin{tabular}[c]{@{}c@{}}5.2\\ $\pm$ \textit{0.0}\end{tabular} &
  \begin{tabular}[c]{@{}c@{}}33.5\\ $\pm$ \textit{0.4}\end{tabular} &
  \begin{tabular}[c]{@{}c@{}}39.6\\ $\pm$ \textit{0.2}\end{tabular} &
  \begin{tabular}[c]{@{}c@{}}2.3\\ $\pm$ \textit{0.0}\end{tabular} &
  \begin{tabular}[c]{@{}c@{}}31.8\\ $\pm$ \textit{0.3}\end{tabular} &
  \begin{tabular}[c]{@{}c@{}}40.1\\ $\pm$ \textit{0.4}\end{tabular} \\
CodeLlama-13b &
  \begin{tabular}[c]{@{}c@{}}3.2\\ $\pm$ \textit{0.0}\end{tabular} &
  \begin{tabular}[c]{@{}c@{}}15.8\\ $\pm$ \textit{0.0}\end{tabular} &
  \begin{tabular}[c]{@{}c@{}}16.2\\ $\pm$ \textit{0.0}\end{tabular} &
  \begin{tabular}[c]{@{}c@{}}2.0\\ $\pm$ \textit{0.0}\end{tabular} &
  \begin{tabular}[c]{@{}c@{}}14.6\\ $\pm$ \textit{0.0}\end{tabular} &
  \begin{tabular}[c]{@{}c@{}}15.1\\ $\pm$ \textit{0.0}\end{tabular} \\ \bottomrule
\end{tabular}
\end{scriptsize}
\caption{Evaluation of the GOLD model on two datasets with no description (n/a), logic-forms (LF), and natural language descriptions (NLD). Both the mean and standard errors of the accuracy metrics are presented.}
\label{tab:nldvslf}
\end{table}

\textcolor{black}{Conversely, when using Llama2-13b-chat (Llama2) and CodeLlama-13b (CodeLlama) as the problem-solving module, we implement instructions to guide the generation of answers. Since their generations differ from the ground truth, we opt to calculate the accuracy of choosing the correct option from given candidates. According to Table~\ref{tab:nldvslf}, using natural language descriptions significantly enhances the accuracy of the Llama2 model compared to using logic forms, demonstrating the greater compatibility of our natural language descriptions with models like Llama2. However, neither natural language descriptions nor logic forms yield satisfactory outcomes with CodeLlama, possibly due to a mismatch between the training corpus of CodeLlama and the description formats.}

\textcolor{black}{Lastly, we conduct experiments by excluding relevant modules used to generate the natural language descriptions and solely inputting the problem text \(\mathcal{T}\) into the problem-solving module. The results in Table~\ref{tab:nldvslf} show a substantial decline in the performance of the GOLD model across all selected LLMs, highlighting the importance of diagram descriptions provided by relevant modules of the GOLD model in solving geometry math problems.}

\subsection{Accuracy of the Extraction of geo2geo and sym2geo Relations}
\label{sec:Ablation-Study-on-feature-embedding-and-spatial-embedding}

\textcolor{black}{Our analysis in Table~\ref{tab:feature-spatial-embedding} and measured by F1 metric, evaluates the accuracy of extracting geometric relations with and without \(\text{emb}_{\mathit{feat}}\) and \(\text{emb}_{\mathit{spat}}\) on PGPS9K test subset. We note that the pre-parsing stage achieves a high F1-score of 98.9\%, ensuring accurate identification of symbols and geometric primitives for \textit{sym2geo} and \textit{geo2geo} relations extraction. However, when directly using \(\mathbf{V}^{s, g}\) as vectors of symbols and geometric primitives (only using feature outputs from the pre-parsing step), the absence of \(\text{emb}_{\mathit{feat}}\) and \(\text{emb}_{\mathit{spat}}\) leads to a notable decrease in performance for both relations extraction. Conversely, the inclusion of either \(\text{emb}_{\mathit{feat}}\) and \(\text{emb}_{\mathit{spat}}\) results in improved performance. Table~\ref{tab:feature-spatial-embedding} further reveals that the extraction of both relation types reaches its highest F1-score when both embeddings are utilized. These results highlight the advantages of our approach in separately modelling symbols and geometric primitives, which proves to be more efficient in addressing the relation extraction of geometry math problems (please see Appendix~\ref{sec:appendix-Influence} for the impact of \(\text{emb}_{\mathit{feat}}\) and \(\text{emb}_{\mathit{spat}}\) on problem-solving accuracy, and Appendix~\ref{sec:appendix-geo-type-embedding} for the ablation analysis for the \(\text{emb}^{g}_{\mathit{type}}\)).}

\begin{table}[tbhp!]
\begin{small}
\centering
\begin{tabular}{@{}ccccc@{}}
\toprule
$\text{emb}_{\mathit{feat}}$ & $\text{emb}_{\mathit{spat}}$ & pre-parsing & geo2geo   & sym2geo  \\ \midrule
          &           & 98.9      & 65.2 $\pm$ \textit{0.1}  & 58.6 $\pm$ \textit{0.1} \\ 
      \faCheck     &           & 98.9      & 79.8 $\pm$ \textit{0.3}  & 75.6 $\pm$ \textit{0.5} \\ 
          &      \faCheck      & 98.9      & 80.6 $\pm$ \textit{0.4} & 71.1 $\pm$ \textit{0.2} \\ 
      \faCheck    &     \faCheck        & 98.9    & \textbf{93.7} $\pm$ \textit{0.2} & \textbf{77.3} $\pm$ \textit{0.1} \\ \bottomrule
\end{tabular}
\caption{The check mark (\faCheck) indicates that the corresponding embedding is enabled. Note that "pre-parsing" is not influenced by $\text{emb}_{\mathit{feat}}$ and $\text{emb}_{\mathit{spat}}$. Both the mean and standard errors of the accuracy metrics are presented. See Appendix~\ref{sec:appendix-image-parsing-accuracy} and~\ref{sec:appendix-relation-parsing-accuracy} for the accuracy of fine-grained relations.}
\label{tab:feature-spatial-embedding}
\end{small}
\end{table}

\begin{figure}[tbhp!]
\centering
\includegraphics[width=1.0\columnwidth]{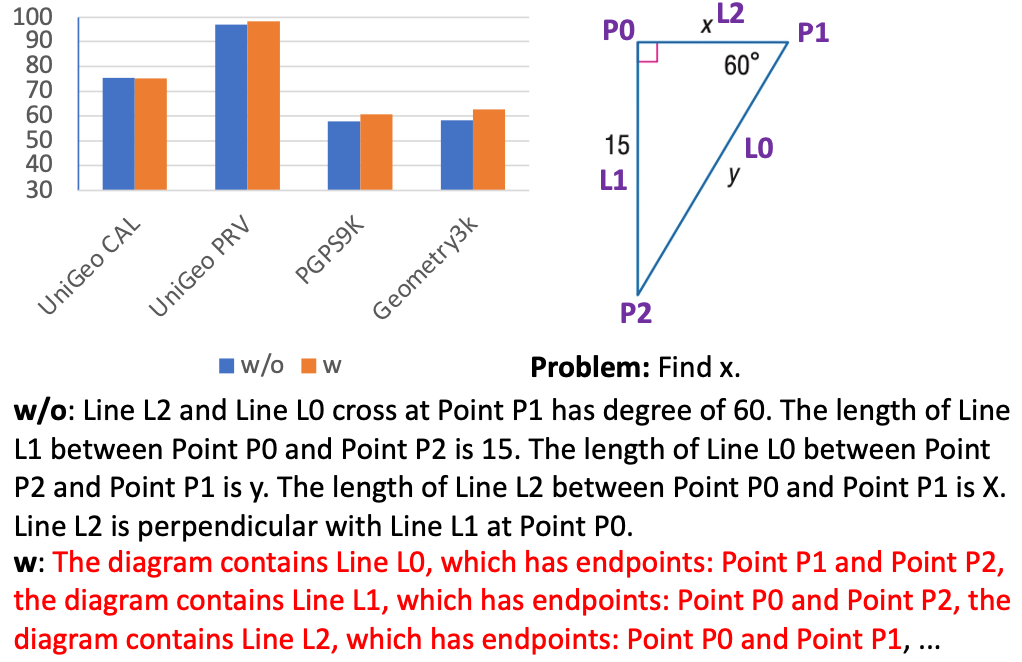} 
\caption{Top-left: the performance of the GOLD (using T5-base) with (\textbf{w}) and without (\textbf{w/o}) the \textit{geo2geo}. Top-right: Geometry math problem. Bottom: Predicted diagram description with and without the \textit{geo2geo}. The same text between (\textbf{w}) and (\textbf{w/o}) is omitted for space consideration, where the red text is \textit{geo2geo} relations.}
\label{fig:geo2geo}
\end{figure}

Table~\ref{tab:feature-spatial-embedding} shows that the GOLD model accurately captures \textit{geo2geo} relation, prompting us to investigate its impact on solving geometry math problems. The bar chart in Figure~\ref{fig:geo2geo} indicates a notable decline in model performance on the PGPS9K and Geometry3K datasets when \textit{geo2geo} relations are omitted. However, this trend is less pronounced on the UniGeo datasets. This is likely because the PGPS9K and Geometry3K datasets often lack descriptions of geometric primitives in their problem texts. An example from the Geometry3K dataset, illustrated in Figure~\ref{fig:geo2geo}, demonstrates this issue: the problem text typically poses a question (e.g., "Find X") without extra information. Consequently, relying only on \textit{sym2geo} relations leads to insufficient representation of essential diagram details.

\section{Conclusion}

In this work, we have introduced the GOLD model for automated geometry math problem-solving. GOLD uniquely converts geometry diagrams into natural language descriptions, facilitating direct integration with LLMs for problem-solving. \textcolor{black}{A key feature of the GOLD model is that it separately handles symbols and geometric primitives, simplifying the process of establishing relations between symbols and geometric primitives and relations among geometric primitives themselves.} Our experiments show that the GOLD model outperforms the Geoformer, the previous SOTA on the UniGeo dataset, with accuracy improvements of 12.7\% and 42.1\% on the UniGeo calculation and proving datasets, respectively. Additionally, compared to PGPSNet, the SOTA for the PGPS9K and Geometry3K datasets, the GOLD model shows notable accuracy improvements of 1.8\% and 3.2\%, respectively, showing our method's effectiveness.



\section{Limitations}

While our GOLD model marks a significant advancement in solving geometry math problems, areas remain for future improvement. For example, the GOLD model has not yet reached the level of human performance in solving geometry math problems. This gap is possibly due to the limitations in fully extracting geometric relations from diagrams. While GOLD accurately identifies symbols, geometric primitives, and \textit{geo2geo} relations, the extraction of \textit{sym2geo} relations still requires enhancement. Moreover, this study evaluated three popular large language models (LLMs): T5-bases, Llama2-13b-chat, and CodeLlama-13b. To deepen our understanding and leverage the full potential of LLMs in solving geometry math problems, it would be beneficial to assess more LLMs. This broader evaluation could provide more comprehensive insights into optimizing LLMs for this specific task.

\bibliography{anthology,custom}
\bibliographystyle{acl_natbib}

\appendix

\section{Inference}
\label{sec:appendix_Inference}

During the inference stage, we employ Eq~\ref{eq:symbol_vector} and Eq~\ref{eq:geometry_primitive_vector} to map symbols \(s \in \mathcal{S}\) and geometric primitives \(g \in \mathcal{G}\) to corresponding vectors \(\text{vec}^{s \in \mathcal{S}}\) and \(\text{vec}^{g \in \mathcal{G}}\), respectively. Following this, we proceed with the inference of \textit{sym2geo} and \textit{geo2geo} relations.

\subsection{Predict sym2geo Relation}
\label{sec:appendix_Inference_sym2geo}

For a text symbol \(s \in \{\mathcal{S} | \text{cls}_s = \text{"text"}\}\), it is necessary to determine its meaning based on its \textit{text\_class}. To accomplish this, we assign the category of text symbol \(s \in \{\mathcal{S} | \text{cls}_s = \text{"text"}\}\) as the one with the highest probability among the \(P(\textit{text\_class}|s)\) values, as specified in Eq~\ref{eq:text-class}:

\begin{equation}
\small
    \mathit{text\_class}_{s} = \textit{argmax} P(\mathit{text\_class}|s)
\end{equation}

\begin{itemize}
    \item if \(\textit{text\_class}_{s}\) is 0 (i.e., category \(\mathit{0}\)), it indicates that the symbol \(s\) corresponds to the reference name of a point, or a line, or a circle. In this case, we assign the symbol \(s\) to the geometric primitive \(g\) that has the highest probability of \(P(\text{rel}^{\mathit{sym2geo}}_{s \in \{\mathcal{S} | \text{cls}_s = \text{"text"}\}, g \in \{\mathbf{P}, \mathbf{L}, \mathbf{C}\}}|s,g)\), where \( g \in \{\mathbf{P}, \mathbf{L}, \mathbf{C}\}\) specifies that the geometric primitive \(g\) belongs to the set of points, lines, and circles:
    \begin{equation}
    \small
        g = \textit{argmax} P(\text{rel}^{\mathit{sym2geo}}_{s \in \{\mathcal{S} | \text{cls}_s = \text{"text"}\}, g \in \{\mathbf{P}, \mathbf{L}, \mathbf{C}\}}|s,g)
    \end{equation}

    \item if \(\textit{text\_class}_{s}\) is 1 (i.e., category \(\mathit{1}\)), it indicates that the symbol  \(s\) represents the degree of an angle. Since an angle consists of two lines and one point, we select the point with the highest probability \(P(\text{rel}^{\mathit{sym2geo}}_{s \in \{\mathcal{S} | \text{cls}_s = \text{"text"}\},g\in\{\mathbf{P}\}}|s,g)\),  and we select the two lines with the top two highest probabilities \(P(\text{rel}^{\mathit{sym2geo}}_{s \in \{\mathcal{S} | \text{cls}_s = \text{"text"}\},g\in\{\mathbf{L}\}}|s,g)\). It is worth mentioning these two lines must have \(geo2geo\) relations of "end-point" or "on-a-line" with the selected point.
    \begin{equation}
    \small
    \begin{aligned}
        & p=\textit{argmax} P(\text{rel}^{\mathit{sym2geo}}_{s \in \{\mathcal{S} | \text{cls}_s = \text{"text"}\},g\in\{\mathbf{P}\}}|s,g) \\
        & l_1,l_2=\textit{argmax}_{two} P(\text{rel}^{\mathit{sym2geo}}_{s \in \{\mathcal{S} | \text{cls}_s = \text{"text"}\},g\in\{\mathbf{L}\}}|s,g), \\
        & \text{where } \mathit{rel}_{l_1, p} \in \{\text{"end-point"`, "on-a-line"}\} \text{ and} \\
        & \mathit{rel}_{l_2, p} \in \{\text{"end-point"`, "on-a-line"}\}
    \end{aligned}
    \end{equation}

    \item if \(\textit{text\_class}_{s}\) is 2 (i.e., category \(\mathit{2}\)), it indicates that the symbol \(s\) represents the length of a line. Since a line consists of two points, we select the points with the top two highest probabilities  \(P(\text{rel}^{\mathit{sym2geo}}_{s \in \{\mathcal{S} | \text{cls}_s = \text{"text"}\},g\in\{\mathbf{P}\}}|s,g)\):
    \begin{equation}
    \small
        p_1, p_2=\textit{argmax}_{two} P(\text{rel}^{\mathit{sym2geo}}_{s \in \{\mathcal{S} | \text{cls}_s = \text{"text"}\},g\in\{\mathbf{P}\}}|s,g)
    \end{equation}

    \item if \(\textit{text\_class}_{s}\) is 3 (i.e., category \(\mathit{3}\)), it indicates that the symbol \(s\) represents the degree of an angle on the circle. In this case, the angle is formed by the centre point of a circle and two points lying on the arc of a circle. Therefore, we first select the circle with the highest probability of \(P(\text{rel}^{\mathit{sym2geo}}_{s \in \{\mathcal{S} | \text{cls}_s = \text{"text"}\},g\in\{\mathbf{C}\}}|s,g)\). Subsequently, we select two points with the top two highest probabilities \(P(\text{rel}^{\mathit{sym2geo}}_{s \in \{\mathcal{S} | \text{cls}_s = \text{"text"}\},g\in\{\mathbf{P}\}}|s,g)\). Worth mentioning, these two points must be on the arc of the selected circle:
    \begin{equation}
    \small
    \begin{aligned}
        & c=\textit{argmax} P(\text{rel}^{\mathit{sym2geo}}_{s \in \{\mathcal{S} | \text{cls}_s = \text{"text"}\},g\in\{\mathbf{C}\}}|s,g) \\
        & p_1, p_2 = \textit{argmax}_{two} P(\text{rel}^{\mathit{sym2geo}}_{s \in \{\mathcal{S} | \text{cls}_s = \text{"text"}\},g\in\{\mathbf{P}\}}|s,g) \text{,} \\
       & \text{where } \mathit{rel}_{p_1, c} =  \mathit{rel}_{p_2, c} = \text{"on-a-circle"} \\
    \end{aligned}
    \end{equation}
\end{itemize}

For the geometric relations among geometric primitives, such as parallel. It is determined by the  \textit{other2geo} relation. For the \textit{other2geo} relation involving other symbols, it is required that the relation holds with at least two geometric primitives. This means that there should be at least two geometric primitives with probabilities \(P(\text{rel}^{\mathit{sym2geo}}_{s, g}|s,g)\) larger than a threshold \(\theta\). In this case, the geometric primitives are selected based on this criterion.

\begin{equation}
\small
\begin{aligned}
   & \{g_{\text{indices}}\}=\text{sorted}(P(\text{rel}^{\mathit{sym2geo}}_{s\in \{\mathcal{S} | \text{cls}_s \neq  \text{"text"}\},g\in{\{\mathbf{P,L,C}}}\}|s,g)) > \theta \\
   & {g_{\text{selected}}} = \mathcal{G}[\{g_{\text{indices}}\}]
\end{aligned}
\end{equation}

where "sorted" indicates that values are sorted in descending order, and \([]\) refers to the selection from the geometric primitives group \(\mathcal{G}\) according to the indices \(\{g_{\text{indices}}\}\). The threshold \(\theta\) is set as 0.5 experimentally.

\subsection{Predict geo2geo Relation}
\label{sec:appendix_Inference_geo2geo}

The \textit{geo2geo} relation between geometric primitives \(g_i\) and \(g_j\) is determined based on Eq~\ref{eq:geo2geo-relation}, where it is assigned as the relation with the highest probability:

\begin{equation}
\small
    \mathit{rel}_{g_i, g_j \in \mathcal{G}} = \textit{argmax}  P(\text{rel}^{\mathit{geo2geo}}_{g_i,g_j}|g_i,g_j)
\end{equation}

In an ideal scenario, the OCR results would accurately provide references to the points, lines, and circles, allowing us to extract precise information about the geometric primitives. However, the open-source OCR tool\footnote{\url{https://github.com/JaidedAI/EasyOCR}} we have adopted is not accurate. As a result, some primitives may lack reference names. To address this issue, we automatically label the primitives in sequential order (e.g., "P1, P2, L1, L2") if their reference names are missing.

\subsection{Generate Solution Program}

Once the \textit{geo2geo} and \textit{sym2geo} relations are constructed, we proceed to convert them into natural language descriptions \(\mathcal{L}\) (See Appendix~\ref{sec:appendix_convert} for details). We then concatenate the natural language descriptions \(\mathcal{L}\) with the problem text \(\mathcal{T}\). This combined text is passed to the problem-solving module, which employs BeamSearch with a beam size of 10 to generate the solution program \(\mathcal{P}\). \textcolor{black}{Moreover, when using larger LLMs, such as Llama2, we add instructions in front of the concatenation of \(\mathcal{L}\) and \(\mathcal{T}\), which is further sent to LLMs to generate reasoning process.}

\section{Convert Relations to Natural Language Descriptions}
\label{sec:appendix_convert}

\begin{table*}[tbht!]
\begin{small}
\centering
\begin{tabular}{@{}clll@{}}
\toprule
 &
  \multicolumn{1}{c}{Relations} &
  \multicolumn{1}{c}{Paradigm} &
  \multicolumn{1}{c}{Example} \\ \midrule
 &
  Point &
  The diagram contains \$\{\}. &
  The diagram contains \textbf{Point} \textbf{A}, \textbf{B}, \textbf{C}. \\ \cmidrule(l){2-4} 
geo2geo &
  Line &
  \begin{tabular}[c]{@{}l@{}}The diagram contains \$\{\}, \\ which has endpoints: \$\{\} and \$\{\}, \\ In addition, there is/are \$\{\} on the line.\end{tabular} &
  \begin{tabular}[c]{@{}l@{}}The diagram contains \textbf{Line L1}, \\ which has endpoints: \textbf{Point P0} and \textbf{Point P1},\\ In addition, there is/are \textbf{Point P2} on the line.\end{tabular} \\ \cmidrule(l){2-4} 
 &
  Circle &
  \begin{tabular}[c]{@{}l@{}}The diagram contains \$\{\},\\ whose center point is \$\{\},\\ which has \$\{\} on its arc. \end{tabular} &
  \begin{tabular}[c]{@{}l@{}}The diagram contains \textbf{Circle M},\\ whose center point is \textbf{Point E},\\ which has \textbf{Point F}, \textbf{Point G} on its arc.\end{tabular} \\ \midrule
 &
  Degree &
  \begin{tabular}[c]{@{}l@{}}1. Angle \$\{\} has degree of \$\{\}.\\ 2. Line \$\{\} and Line \$\{\} cross at Point \$\{\}\\     has degree of \$\{\}.\end{tabular} &
  \begin{tabular}[c]{@{}l@{}}1. Angle \textbf{1} has degree of \textbf{100}.\\ 2. Line \textbf{L1} and Line \textbf{L2} cross at Point \textbf{C}\\     has degree of \textbf{50}.\end{tabular} \\ \cmidrule(l){2-4} 
text2geo &
  Length &
  \begin{tabular}[c]{@{}l@{}}The length of Line \$\{\} between Point \$\{\}\\ and Point \${} is \$\{\}.\end{tabular} &
  \begin{tabular}[c]{@{}l@{}}The length of Line \textbf{L3} between Point \textbf{A}\\ and Point \textbf{B} is \textbf{10}.\end{tabular} \\ \cmidrule(l){2-4} 
 &
  Circle Degree &
  \begin{tabular}[c]{@{}l@{}}Line \$\{\} and Line \$\{\} cross at the \\ center point \$\{\} of Circle \$\{\} has\\ degree of \$\{\}.\end{tabular} &
  \begin{tabular}[c]{@{}l@{}}Line \textbf{L1} and Line \textbf{L2} cross at the\\ center point \textbf{C} of Circle \textbf{C0} has\\ degree of \textbf{20}.\end{tabular} \\ \midrule
 &
  same degree &
  \begin{tabular}[c]{@{}l@{}}Angle \$\{\} has the same degree with\\ Angle \$\{\} ...\end{tabular} &
  \begin{tabular}[c]{@{}l@{}}Angle \textbf{1} has the same degree with\\ Angle \textbf{2}, Angle \textbf{3}.\end{tabular} \\ \cmidrule(l){2-4} 
other2geo &
  same length &
  \begin{tabular}[c]{@{}l@{}}Line \$\{\} has the same length with\\ Line \$\{\} ...\end{tabular} &
  \begin{tabular}[c]{@{}l@{}}Line \textbf{L1} has the same length with\\ Line \textbf{L2}, Line \textbf{L3}.\end{tabular} \\ \cmidrule(l){2-4} 
 &
  parallel &
  Line \$\{\} is parallel with Line \$\{\}... &
  Line \textbf{a} is parallel with Line \textbf{b}. \\ \cmidrule(l){2-4} 
 &
  perpendicular &
  \begin{tabular}[c]{@{}l@{}}Line \$\{\} is perpendicular with Line \$\{\}\\ at Point \$\{\}.\end{tabular} &
  \begin{tabular}[c]{@{}l@{}}Line \textbf{L1} is perpendicular with Line \textbf{L2}\\ at Point \textbf{C}.\end{tabular} \\ \bottomrule
\end{tabular}
\caption{The defined paradigm used to convert \textit{geo2geo} and \textit{sym2geo} relations to natural language descriptions \(\mathcal{L}\). "\$\{\}" is the placeholder. The placeholder is filled in as demonstrated in the "Example" column, and the filled content is highlighted in bold type.}
\label{tab:paradigm}
\end{small}
\end{table*}

Once the \textit{geo2geo} relations and \textit{sym2geo} relations have been established, we proceed to convert these relations into natural language descriptions denoted as \(\mathcal{L}\) following the guidelines specified in Table~\ref{tab:paradigm}.

To begin, we initiate the process by representing the existing geometric primitives in the diagram by enumerating points, lines, and circles within the description of the \textit{geo2geo} relation. In detail, we sequentially enumerate all existing points, providing their reference names as described in the "Point" entry of Table~\ref{tab:paradigm}. We describe the associated points for each line by mentioning their reference names. Additionally, we include a list of points that have "end-point" and "on-a-line" relations with the line, as specified in the "Line" entry of Table~\ref{tab:paradigm}. Similarly, for each circle, we mention its reference name and proceed to list the points that exhibit "center-point" and "on-a-circle" relations with the circle, following the guidelines provided in the "Circle" entry of Table~\ref{tab:paradigm}.

Next, we proceed to describe the \textit{text2geo} relation within the \textit{sym2geo} relation based on the predicted \textit{text\_class}. Here are the guidelines for each case:
\begin{itemize}
    \item If the \textit{text\_class} indicates that the symbol refers to the reference name of a point (or a line, or a circle), we modify the name of the corresponding point (or line, or circle) accordingly.
    \item If the \textit{text\_class} indicates that the symbol refers to the degree of an angle, we describe it following the guidelines specified in the "Degree" entry of Table~\ref{tab:paradigm}.
    \item If the \textit{text\_class} indicates that the symbol refers to the length of a line, we describe it according to the instructions provided in the "Length" entry of Table~\ref{tab:paradigm}.
    \item If the \textit{text\_class} indicates that the symbol refers to the degree of an angle on the circle, we describe it based on the guidelines outlined in the "Circle Degree" entry of Table~\ref{tab:paradigm}.
\end{itemize}

Furthermore, when dealing with the \textit{other2geo} relations, we describe them based on the specific type of geometric relation as indicated in Table~\ref{tab:paradigm}.

\section{Instruction Choice}
\label{sec:appendix_instruciton_choice}

\textcolor{black}{Instructions serve as direct and explicit commands that clearly communicate to the model the specific task it is required to perform. For our experiments, we initially selected two distinct instruction templates for Llama2-13b-chat \cite{llama2} and CodeLlama-13b \cite{codellama}, as detailed in Table~\ref{tab:instruction-templates}. Upon experimental evaluation, it was observed that the instruction template modified from the one used to train the Llama2 model (displayed at the upper row in Table~\ref{tab:instruction-templates}) demonstrated superior performance. Consequently, we opted for this template in our work.}

\begin{table*}[tbpt!]
\centering
\begin{scriptsize}
\begin{tabular}{@{}ll@{}}
\toprule
Instruction Template &
  Example \\ \midrule
\begin{tabular}[c]{@{}l@{}}{[}INST{]}  \\ \\ You are a problem-solving bot, \\ and now I ask you to solve a geometry problem, \\ please answer the question and provide the correct option letter. \\ The problem is as follows:\\ \\ \{Problem Text\}\\ \\ Here are the basic descriptions of the diagram:\\ \\ \{Natural Language Descriptions\}\\ \\ The Answer and the Reason Process are:\\ \\ {[}/INST{]}\end{tabular} &
  \begin{tabular}[c]{@{}l@{}}{[}INST{]} \\ \\ You are a problem-solving bot, \\ and now I ask you to solve a geometry problem, \\ please answer the question and provide the correct option letter. \\ The problem is as follows:\\ \\ Find the perimeter of the polygon. \\ The Choices are:  A: 20.0, B: 24.0, C: 28.0, D: 34.409,\\ \\ Here are the basic description of the diagram:  \\ \\ The diagram contains Point P0, Point P1, Point P2, Point P3, Point P4, \\ The diagram contains Line L0, which has endpoints: Point P1, Point P3, \\ Line L1, which has endpoints: Point P1, Point P4, \\ Line L2, which has endpoints: Point P3, Point P4, \\ Line L3, which has endpoints: Point P0, Point P3, \\ Line L4, which has endpoints: Point P0, Point P1, \\ Line L5, which has endpoints: Point P0, Point P4, \\ The length of Line L0 between Point P2 and Point P3 is 7.\\ The length of Line L4 between Point P2 and Point P1 is 7.\\ The length of Line L5 between Point P4 and Point P2 is 5.\\ Line L3 between Point P0 and Point P3 has the same length \\ with Line L4 between Point P1 and Point P0 \\ and Line L2 between Point P3 and Point P4 \\ and Line L1 between Point P1 and Point P4. \\ \\ The Answer and the Reason Process are: \\ \\ {[}/INST{]}\end{tabular} \\ \midrule
\begin{tabular}[c]{@{}l@{}}Hint: Please answer the question and provide the correct option letter, \\ e.g., A, B, C, D, at the end\\ \\ \{Problem Text\}\\ \\ Here are the basic descriptions of the diagram:\\ \\ \{Natural Language Descriptions\}\end{tabular} &
  \begin{tabular}[c]{@{}l@{}}Hint: Please answer the question and provide the correct option letter, \\ e.g., A, B, C, D, at the end\\ \\ Find the perimeter of the polygon. \\ The Choices are:  A: 20.0, B: 24.0, C: 28.0, D: 34.409,\\ \\ Here are the basic descriptions of the diagram:\\ \\ The diagram contains Point P0, Point P1, Point P2, Point P3, Point P4, \\ The diagram contains Line L0, which has endpoints: Point P1, Point P3, \\ Line L1, which has endpoints: Point P1, Point P4, \\ Line L2, which has endpoints: Point P3, Point P4, \\ Line L3, which has endpoints: Point P0, Point P3, \\ Line L4, which has endpoints: Point P0, Point P1, \\ Line L5, which has endpoints: Point P0, Point P4, \\ The length of Line L0 between Point P2 and Point P3 is 7.\\ The length of Line L4 between Point P2 and Point P1 is 7.\\ The length of Line L5 between Point P4 and Point P2 is 5.\\ Line L3 between Point P0 and Point P3 has the same length \\ with Line L4 between Point P1 and Point P0 \\ and Line L2 between Point P3 and Point P4 \\ and Line L1 between Point P1 and Point P4.\end{tabular} \\ \bottomrule
\end{tabular}
\end{scriptsize}
\caption{Two instruction templates. The template in the upper row is modified from the instruction used to train the Llama2 model, and another one is from the \citealp{MathVista}. In the column of  "Instruction Template", the "\{problem Text\}" is the geometry math problem text \(\mathcal{T}\), and "\{Natural Language Descriptions\}" is the description of the diagram \(\mathcal{L}\).}
\label{tab:instruction-templates}
\end{table*}

\section{Loss Function Details}
\label{sec:appendix-loss-function-details}

\textcolor{black}{The \({L}_{g2g}\) is defined as the negative log-likelihood loss, where we aim to minimize the negative log-likelihood of the ground truth relations among geometric primitives:}
\begin{equation}
\small
    \mathit{L}_{\mathit{g2g}} = - \sum_{g_i \in \mathbf{P}} \sum_{g_j \in \mathbf{L, C}} \text{log} (P(\text{rel}^{\mathit{geo2geo}}_{g_i,g_j}|g_i,g_j))
\end{equation}
\textcolor{black}{where \(g_i\) is a geometric primitive belonging to points, and \(g_j\) is a geometric primitive belonging to lines and circles. The \(\text{rel}^{\mathit{geo2geo}}_{g_i,g_j}\) refers to the ground truth relation between \(g_i\) and \(g_j\).}

\textcolor{black}{The \({L}_{\mathit{t\_cls}}\) is defined as the negative log-likelihood loss, where we aim to minimize the negative log-likelihood of the ground truth \textit{text\_class} of the text symbol:}
\begin{equation}
\small
    \mathit{L}_{\mathit{t\_cls}}\mathbin{=} \mathbin{-} \sum_{S} \text{log} (P(\textit{text\_class}_{s}|s))
\end{equation}
\textcolor{black}{where \(\text{text\_class}_s\) is the ground truth \(text\_class\) of the symbol \(s\).}

\textcolor{black}{The \(\mathit{L}_{\mathit{s2g}}\) is the binary cross-entropy loss:}
\begin{equation}  
\small
\begin{aligned}
 \mathit{L}_{\mathit{s2g}} \mathbin{=} & \mathbin{-} \sum_{s \in \mathcal{S}} \sum_{g \in \mathcal{G}} \{ \mathbb{I}(s, g) \times \text{log} (P(\text{rel}^{\mathit{sym2geo}}_{s,g}|s,g))  \\ 
 & \mathbin{+} (1 \mathbin{-} \mathbb{I}(s, g)) \times (1 \mathbin{-} \text{log} (P(\text{rel}^{\mathit{sym2geo}}_{s,g}|s,g) \}
\end{aligned}
\end{equation}
\textcolor{black}{where \(\mathbb{I}(s, g)\) is 1 if there is relation between symbol \(s\) and geometric primitive \(g\), otherwise it is 0.}

\textcolor{black}{The \( \mathit{L}_{\mathit{prog}}\) is defined as the negative log-likelihood loss, where we aim to minimize the negative log-likelihood of the tokens of the ground truth solution programs:}
\begin{equation}
\small
    \mathit{L}_{\mathit{prog}}\mathbin{=} \mathbin{-} \sum_{i} \text{log} (P(t_i|t_{<i}))
\end{equation}
\textcolor{black}{where \(i\) is the \(i\)-th token in the ground truth solution program.}

\section{Image Parsing Accuracy}
\label{sec:appendix-image-parsing-accuracy}

\begin{table}[tbht!]
\begin{small}
\centering
\begin{tabular}{@{}lc@{}}
\toprule
  Geometric Primitives or Symbols     & F1 (\%)   \\ \midrule
point  & 99.8 \\
line   & 99.5 \\
circle & 99.1 \\
symbol & 97.2 \\ \bottomrule
\end{tabular}
\caption{Pre-parsing performances by F1 metric.}
\label{tab:image-parsing}
\end{small}
\end{table}

Table~\ref{tab:image-parsing} presents the performance of the image-parsing module, measured using the F1 metric. For geometric primitives, we employ the parsing position evaluation method, utilizing the Hough transform with a distance threshold of 15. For symbols, we use an Intersection over Union (IoU) threshold of 0.5. The results in Table~\ref{tab:image-parsing} demonstrate that the image-parsing module delivers accurate parsing results for diagrams, providing the model with precise information.

\section{Relation Prediction Accuracy}
\label{sec:appendix-relation-parsing-accuracy}

\begin{table}[tbht!]
\begin{small}
\centering
\begin{tabular}{@{}llc@{}}
\toprule
\multicolumn{2}{l}{Relation Type}                              & PGPS9K Test (\%) \\ \midrule
                         & end-point      & 97.9 $\pm$ \textit{0.3}                          \\
\multirow{2}{*}{geo2geo} & on-a-line        & 91.3 $\pm$ \textit{0.4}                            \\
                         & center-point        & 93.6 $\pm$ \textit{0.2}                           \\
                         & on-a-circle      & 92.0 $\pm$ \textit{0.0}                           \\ \midrule
                         & text symbol   & 65.2 $\pm$ \textit{0.1}                         \\
                         & angle         & 73.1 $\pm$ \textit{0.0}                            \\
sym2geo                  & bar           & 75.7 $\pm$ \textit{0.2}                            \\
                         & parallel      & 89.0 $\pm$ \textit{0.4}                            \\
                         & perpendicular & 82.9 $\pm$ \textit{0.0}                            \\ \bottomrule
\end{tabular}
\caption{Relation Parsing performances by F1 metric. Both the mean and standard errors of the accuracy metrics are presented.}
\label{tab:relation-accuracy}
\end{small}
\end{table}

Table~\ref{tab:relation-accuracy} displays the F1 metric for the performance of relation parsing. The results show that our GOLD model accurately predicts \textit{geo2geo} relations. However, for \textit{sym2geo} relations, except for the "parallel" relation, there is considerable room for improvement in the prediction performance.

\section{Influence of feature\_embedding and spatial\_embedding on Geometry Problem Solving}
\label{sec:appendix-Influence}

\textcolor{black}{We conduct ablation study on \textit{feature\_embedding} and \textit{spatial\_embedding} in Table~\ref{tab:feature-spatial-embedding-problem-solving}. To discard the use of (\(\text{emb}_{\mathit{feat}}\) and \(\text{emb}_{\mathit{spat}}\)), we directly use feature outputs from the pre-parsing step as vectors of symbols and geometric primitives, i.e., \(\mathbf{V}^{s, g}\), to construct the \textit{sym2geo} and \textit{geo2geo} relations. We can observe that the GOLD model without any embedding performs the worst on all test subsets. However, when either one of embeddings (\(\text{emb}_{\mathit{feat}}\) or \(\text{emb}_{\mathit{spat}}\)) is added, the model's performance improves. Notably, the model equipped with both embeddings achieves the best performance.}

\begin{table}[tbhp!]
\begin{scriptsize}
\centering
\begin{tabular}{@{}cccccc@{}}
\toprule
$\text{emb}_{\mathit{feat}}$  &
  $\text{emb}_{\mathit{spat}}$ &
  CAL &
  PRV &
  PGPS9K &
  Geometry3K \\ \midrule
 & 
   &
  \begin{tabular}[c]{@{}c@{}}66.2\\ $\pm$ \textit{0.3}\end{tabular} &
  \begin{tabular}[c]{@{}c@{}}90.2\\ $\pm$ \textit{0.2}\end{tabular} &
  \begin{tabular}[c]{@{}c@{}}48.2\\ $\pm$ \textit{0.5}\end{tabular} &
  \begin{tabular}[c]{@{}c@{}}50.2\\ $\pm$ \textit{0.3}\end{tabular} \\ \midrule
 \faCheck &
   &
  \begin{tabular}[c]{@{}c@{}}71.5\\ $\pm$ \textit{0.3}\end{tabular} &
  \begin{tabular}[c]{@{}c@{}}93.2\\ $\pm$ \textit{0.4}\end{tabular} &
  \begin{tabular}[c]{@{}c@{}}55.0\\ $\pm$ \textit{0.1}\end{tabular} &
  \begin{tabular}[c]{@{}c@{}}58.1\\ $\pm$ \textit{0.1}\end{tabular} \\ \midrule
 & \faCheck
   &
  \begin{tabular}[c]{@{}c@{}}72.8\\ $\pm$ \textit{0.2}\end{tabular} &
  \begin{tabular}[c]{@{}c@{}}93.0\\ $\pm$ \textit{0.3}\end{tabular} &
  \begin{tabular}[c]{@{}c@{}}56.3\\ $\pm$ \textit{0.1}\end{tabular} &
  \begin{tabular}[c]{@{}c@{}}58.0\\ $\pm$ \textit{0.2}\end{tabular} \\ \midrule
 \faCheck & \faCheck
   &
 \begin{tabular}[c]{@{}c@{}}\textbf{75.2}\\ $\pm$ \textit{0.3}\end{tabular} &
  \begin{tabular}[c]{@{}c@{}}\textbf{98.5}\\ $\pm$ \textit{0.5}\end{tabular} &
  \begin{tabular}[c]{@{}c@{}}\textbf{60.6}\\ $\pm$ \textit{0.3}\end{tabular} &
  \begin{tabular}[c]{@{}c@{}}\textbf{62.7}\\ $\pm$ \textit{0.2}\end{tabular} \\ \bottomrule
\end{tabular}
\caption{Program accuracy with or without \textit{feature\_embedding} and \textit{spatial\_embedding}. The check mark (\faCheck) indicates that the corresponding embedding is enabled. T5-base is used as the problem-solving module for the GOLD model. Both the mean and standard errors of the accuracy metrics are presented.}
\label{tab:feature-spatial-embedding-problem-solving}
\end{scriptsize}
\end{table}

\begin{figure}[tbht!]
\centering
\includegraphics[width=0.9\columnwidth]{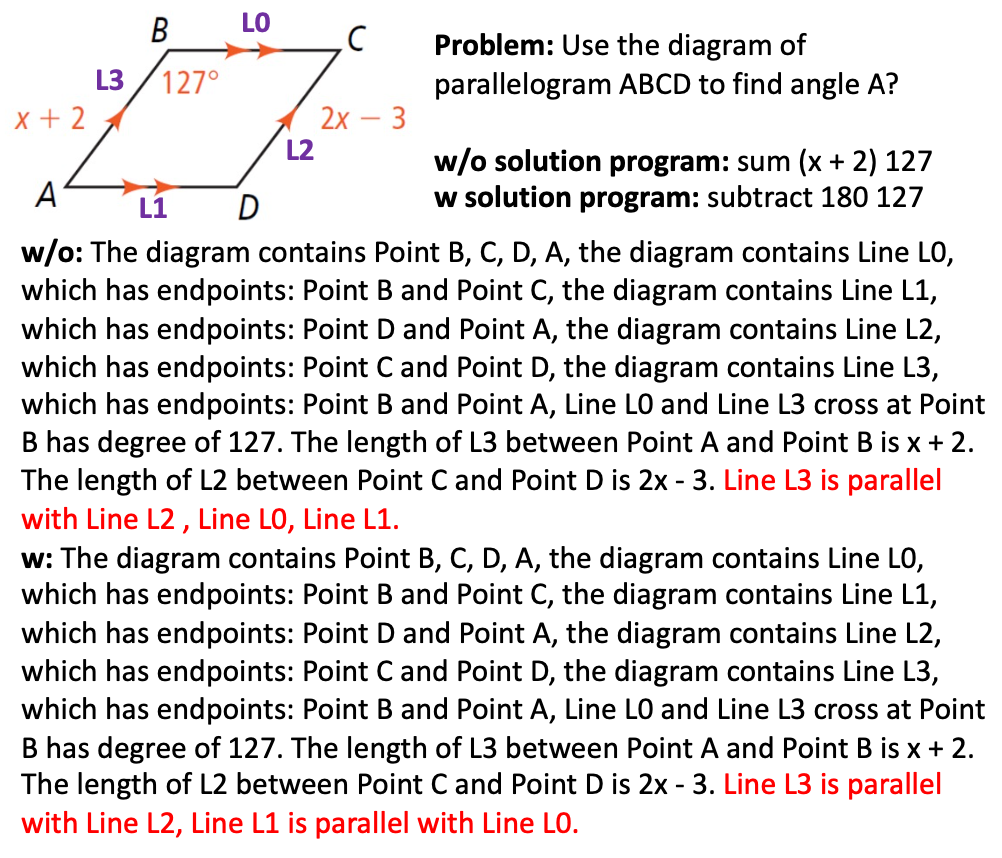} 
\caption{An example from the \textit{111}-th problem in the PGPS9K dataset. This case shows that models' natural language descriptions and solution programs outputs with and without \textit{spatial\_embedding}. The purple notations in the diagram are added by us. Note that the different parts of diagram descriptions between \textbf{w/o} and \textbf{w} are coloured red.}
\label{fig:spatial-embedding}
\end{figure}

\begin{figure}[tbht!]
\centering
\includegraphics[width=0.85\columnwidth]{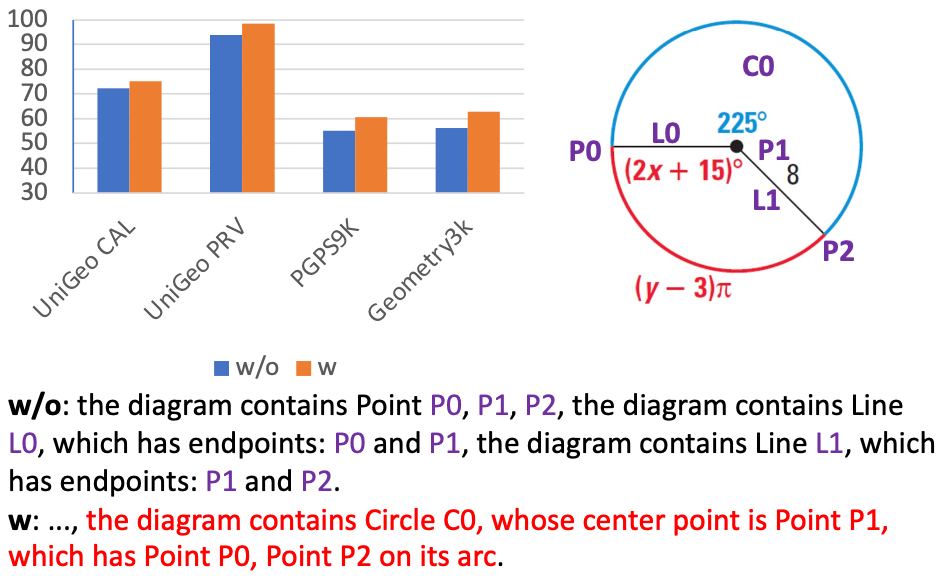} 
\caption{The top-left bar chart compares GOLD (T5-base as the problem-solving module) accuracy in solving geometry math problems, with (\textbf{w}) and without (\textbf{w/o}) the use of \textit{geo\_type\_embedding}. The top-right diagram is from the 375th problem in the PGPS9K dataset, while the bottom part shows the predicted diagram descriptions for two different cases. Purple notations in the diagram are added for better visual comprehension. The differences between the two diagram description texts are highlighted in red. It should be noted that the same texts in the \textbf{w} to the \textbf{w/o} section are omitted, which are represented by "...".}
\label{fig:geo-type}
\end{figure}

In Figure~\ref{fig:spatial-embedding}, we conduct a case study on the GOLD model with and without the use of \textit{spatial\_embedding}. It is evident that the model without \textit{spatial\_embedding} incorrectly generates the "parallel" relation between lines, resulting in an erroneous solution program. This highlights the importance of \textit{spatial\_embedding} in capturing accurate spatial relations and improving the model's performance.

\section{Importance of the geo\_type\_embedding}
\label{sec:appendix-geo-type-embedding}

We conducted experiments to assess the impact of \textit{geo\_type\_embedding} (\(\text{emb}^{g}_{\mathit{type}}\)). The top-left bar chart in Figure~\ref{fig:geo-type} demonstrates that the model's performance declines when \(\text{emb}^{g}_{type}\) is not utilized. Notably, the performance gaps between the model with \(\text{emb}^{g}_{\mathit{type}}\) and without it are more pronounced on the PGPS9K and Geometry3K datasets compared to the UniGeo datasets. We believe this is because the problem text in the UniGeo dataset explicitly mentions the geometric primitives, providing valuable information that helps the GOLD model understand the geometric primitives more effectively. Furthermore, as shown in Figure~\ref{fig:geo-type}, the GOLD model without \(\text{emb}^{g}_{\mathit{type}}\) fails to generate accurate circle information, impeding its ability to further generate correct solution programs.

\end{document}